\documentclass[acmtog, authorversion]{acmart}

\AtBeginDocument{%
  \providecommand\BibTeX{{%
    \normalfont B\kern-0.5em{\scshape i\kern-0.25em b}\kern-0.8em\TeX}}}

\usepackage{bm}
\usepackage{comment}
\usepackage{multirow}
\usepackage{graphicx}
\usepackage{overpic}

\setcopyright{rightsretained}
\copyrightyear{2022}
\acmYear{2022}
\acmDOI{10.1145/3528233.3530740} 

\acmConference[SIGGRAPH '22 Conference Proceedings]{Special Interest Group on Computer Graphics and Interactive Techniques Conference Proceedings}{August 7--11, 2022}{Vancouver, BC, Canada} \acmBooktitle{Special Interest Group on Computer Graphics and Interactive Techniques Conference Proceedings (SIGGRAPH '22 Conference Proceedings), August 7--11, 2022, Vancouver, BC, Canada} 
\acmISBN{978-1-4503-9337-9/22/08}

\begin{document}

\title{Drivable Volumetric Avatars using Texel-Aligned Features}

\author{Edoardo Remelli}
\email{edoremelli@fb.com}
\affiliation{%
  \institution{Meta Reality Labs}
  \country{Switzerland}
}

\author{Timur Bagautdinov}
\email{timurb@fb.com}
\affiliation{%
  \institution{Meta Reality Labs}
  \country{USA}
}

\author{Shunsuke Saito}
\email{shunsukesaito@fb.com}
\affiliation{%
  \institution{Meta Reality Labs}
  \country{USA}
}

\author{Tomas Simon}
\email{tsimon@fb.com}
\affiliation{%
  \institution{Meta Reality Labs}
  \country{USA}
}

\author{Chenglei Wu}
\email{chenglei@fb.com}
\affiliation{%
  \institution{Meta Reality Labs}
  \country{USA}
}

\author{Shih-En Wei}
\email{swei@fb.com}
\affiliation{%
  \institution{Meta Reality Labs}
  \country{USA}
}

\author{Kaiwen Guo}
\email{kwguo@fb.com}
\affiliation{%
  \institution{Meta Reality Labs}
  \country{USA}
}

\author{Zhe Cao}
\email{zhecao@fb.com}
\affiliation{%
  \institution{Meta Reality Labs}
  \country{USA}
}

\author{Fabian Prada}
\email{fabianprada@fb.com}
\affiliation{%
  \institution{Meta Reality Labs}
  \country{USA}
}

\author{Jason Saragih}
\email{jsaragih@fb.com}
\affiliation{%
  \institution{Meta Reality Labs}
  \country{USA}
}

\author{Yaser Sheikh}
\email{yasers@fb.com}
\affiliation{%
  \institution{Meta Reality Labs}
  \country{USA}
}

\renewcommand{\shortauthors}{E. Remelli et al.}

\begin{teaserfigure}
    \centering
    %trim={<left> <lower> <right> <upper>}
    \includegraphics[width=\linewidth]{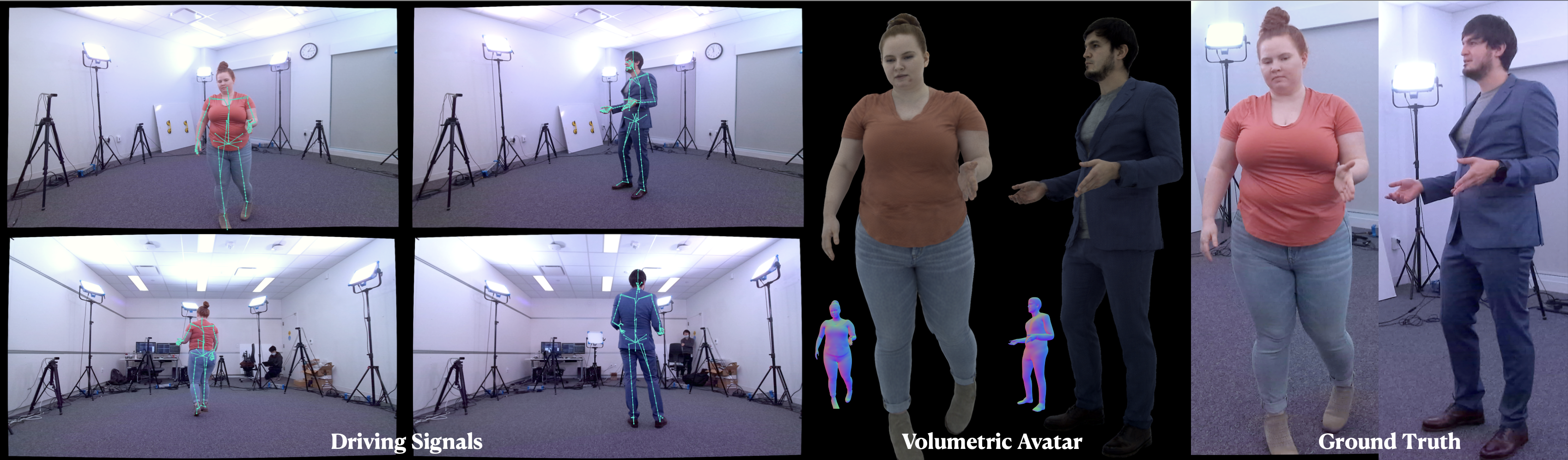}
    \caption{We propose a drivable volumetric model for full-body avatars, 
    which relies on texel-aligned features to be fully faithful to the driving signal.}
    \label{fig:teaser}
\end{teaserfigure}

\begin{abstract}
  A clear and well-documented \LaTeX\ document is presented as an
  article formatted for publication by ACM in a conference proceedings
  or journal publication. Based on the ``acmart'' document class, this
  article presents and explains many of the common variations, as well
  as many of the formatting elements an author may use in the
  preparation of the documentation of their work.
\end{abstract}



\begin{abstract}

Photorealistic telepresence requires both high-fidelity body modeling
\textit{and} faithful driving to enable dynamically synthesized
appearance that is indistinguishable from reality. 
In this work, we propose an end-to-end framework that addresses two core 
challenges in modeling and driving full-body avatars of real people. 
One challenge is driving an avatar while staying faithful to details and 
dynamics that cannot be captured by a global low-dimensional parameterization 
such as body pose. 
Our approach supports driving of clothed avatars with 
wrinkles and motion that a real driving performer exhibits beyond the training 
corpus. 
Unlike existing global state representations or non-parametric screen-space 
approaches, we introduce texel-aligned features---a localised representation 
which can leverage both the structural prior of a skeleton-based parametric model and 
observed sparse image signals at the same time. 
Another challenge is modeling a temporally coherent clothed avatar, which typically requires precise surface tracking. To circumvent this, we propose a novel volumetric avatar representation by extending mixtures of volumetric primitives to articulated objects. By explicitly incorporating articulation, our approach naturally generalizes to unseen poses. We also introduce a localized viewpoint conditioning, which leads to a large improvement in generalization of view-dependent appearance. 
The proposed volumetric representation does not require high-quality mesh tracking as a prerequisite and brings significant quality improvements compared to mesh-based counterparts. 
In our experiments, we carefully examine our design choices and demonstrate the efficacy of our approach, outperforming the state-of-the-art methods on challenging driving scenarios.
\end{abstract}

% One challenge is driving an avatar while staying faithful to details and 
% dynamics that cannot be captured by a global low-dimensional parameterization 
% such as body pose. 
% %
% Clothing state, for example, is not a deterministic function of instantaneous pose, and must therefore be simulated or hallucinated by methods relying on low-dimensional parameterizations. 
% %
% Unlike existing global state representations or non-parametric screen-space 
% approaches, we introduce texel-aligned features---a localised representation 
% which can leverage both the structural prior of a skeleton-based parametric model and 
% observed sparse image signals at the same time. 
% %
% Our approach supports driving of clothed avatars with 
% wrinkles and motion that a real driving performer exhibits beyond the training 
% corpus. 

% various defs
\newcommand{\bb}{\mathbf{b}}
\newcommand{\bx}{\mathbf{x}}
\newcommand{\by}{\mathbf{y}}
\newcommand{\bz}{\mathbf{z}}
\newcommand{\bc}{\mathbf{c}}
\newcommand{\bbf}{\mathbf{f}}
\newcommand{\bv}{\mathbf{v}}
\newcommand{\be}{\mathbf{e}}
\newcommand{\bt}{\mathbf{t}}
\newcommand{\bn}{\mathbf{n}}

\newcommand{\bT}{\mathbf{T}}
\newcommand{\bG}{\mathbf{G}}
\newcommand{\bW}{\mathbf{W}}
\newcommand{\bR}{\mathbf{R}}
\newcommand{\bM}{\mathbf{M}}
\newcommand{\bI}{\mathbf{I}}

\newcommand{\balpha}{\bm{\alpha}}
\newcommand{\bbeta}{\bm{\beta}}
\newcommand{\bgamma}{\bm{\gamma}}
\newcommand{\bdelta}{\bm{\delta}}
\newcommand{\btheta}{\bm{\theta}}

\newcommand{\bmu}{\bm{\mu}}
\newcommand{\bsigma}{\bm{\sigma}}

% bR?
\newcommand{\mR}{\mathbb{R}}
\newcommand{\mL}{\mathcal{L}}
\newcommand{\mN}{\mathcal{N}}

\newcommand{\wrt}{w.r.t.\@\xspace}

\newcommand{\ttt}[1]{\texttt{#1}}


\maketitle

\section{Introduction}

Augmented reality (AR) and virtual reality (VR) have the potential to 
become major computing platforms, enabling people to interact 
with each other in ever more immersive ways across space and time. 
Among these possibilities, authentic social telepresence aims
at life-like presence in AR and VR which is indistinguishable 
from reality.
This imposes a fundamental requirement for techniques to faithfully
teleport every possible detail expressed by humans in reality.   

One promising path to achieve this is to rely on a photorealistic 
animatable model, often obtained with an elaborate capture system, 
which essentially acts as a strong data-driven 
prior~\cite{lombardi2018deep, bagautdinov2021driving, xiang2021modeling}.
Although these methods are capable of producing realistically-looking
free-viewpoint renders, and are robust to occlusions and driving signal
incompleteness, these methods do not fully exploit available inputs.
In practice, such methods typically map dense sensory inputs to
sparse driving signals, such as body pose or low-dimensional embeddings.
Therefore, a large proportion of detailed observations about the subject
are effectively thrown away, resulting in the need to re-hallucinate 
these details in the final render. 
This creates a clear fidelity gap between teleportation and reality, 
resulting in a loss in quality of the conveyed social cues.
One of the reasons why relying exclusively on such model-based methods 
is insufficient lies in the fact that it is non-trivial to design a driving 
representation which is simultaneously expressive and relatively
agnostic to the capture setup to ensure generalization in novel 
conditions. 

An alternative path to building telepresence systems showing
promise, is to rely on model-free methods which combine classical geometry 
reconstruction methods with image-space processing, either with ad-hoc image
fusion~\cite{lawrence2021project} or neural 
re-rendering~\cite{martin2018lookingood, shysheya2019textured}.
% FIXME
Such methods are able to better exploit available inputs,
but usually require highly specialized hardware for high-fidelity 
real-time sensing and would be limited in handling occlusions and 
incomplete data, either exhibiting limited quality rendering of novel 
viewpoints~\cite{martin2018lookingood} or requiring high viewpoint coverage by 
using multiple highly-specialized sensors~\cite{lawrence2021project}. 
Moreover, due to the challenges in real-time sensing, artifacts may 
occur around occlusion boundaries and body parts with limited resolution, 
like fingers or hair.
%
% Similarly, methods that aim at combining low-fidelity model-based
% representations with model-free image-space neural rendering, 
% such as SMPLPix~\cite{prokudin2021}, tend to generalize poorly to novel viewpoints.
%FIXME: Why is this true?

In this work, our goal is to design a method that effectively 
combines the expressiveness of model-free methods with the robustness 
of model-based neural rendering. 
%
%The key idea is to provide the model with a dense conditioning signal, 
%while still relying on data-driven neural rendering model. 
The key idea is a localized state representation, which we call {\em texel-aligned features}, 
that provides the model with a dense conditioning signal while still relying on a 
data-driven neural rendering model. 
%
%Dense conditioning allows us to minimize the amount of information loss, and the data-driven model acts as a strong prior that allows the model to perform well even in scenarios with impoverished sensory input.
Dense conditioning allows us to maximize the amount of information
extracted from the driving signals,
while the data-driven model acts as a strong prior that allows the
model to perform well even in scenarios with impoverished sensory input.
Additionally, our approach uses a hybrid volumetric 
representation specifically tailored to modeling human bodies, 
which exhibits both good generalization to novel poses and
produces high-quality free-viewpoint renders.
This is in contrast to image-space neural rendering methods~\cite{martin2018lookingood} 
and mesh-based~\cite{bagautdinov2021driving} methods, which either
lead to poor generalization on novel views, or are not capable 
of modeling complex geometries with varying topology, which are
abundant in clothed dynamic humans.

\begin{figure*}[t!]
\centering
\includegraphics[width=\textwidth]{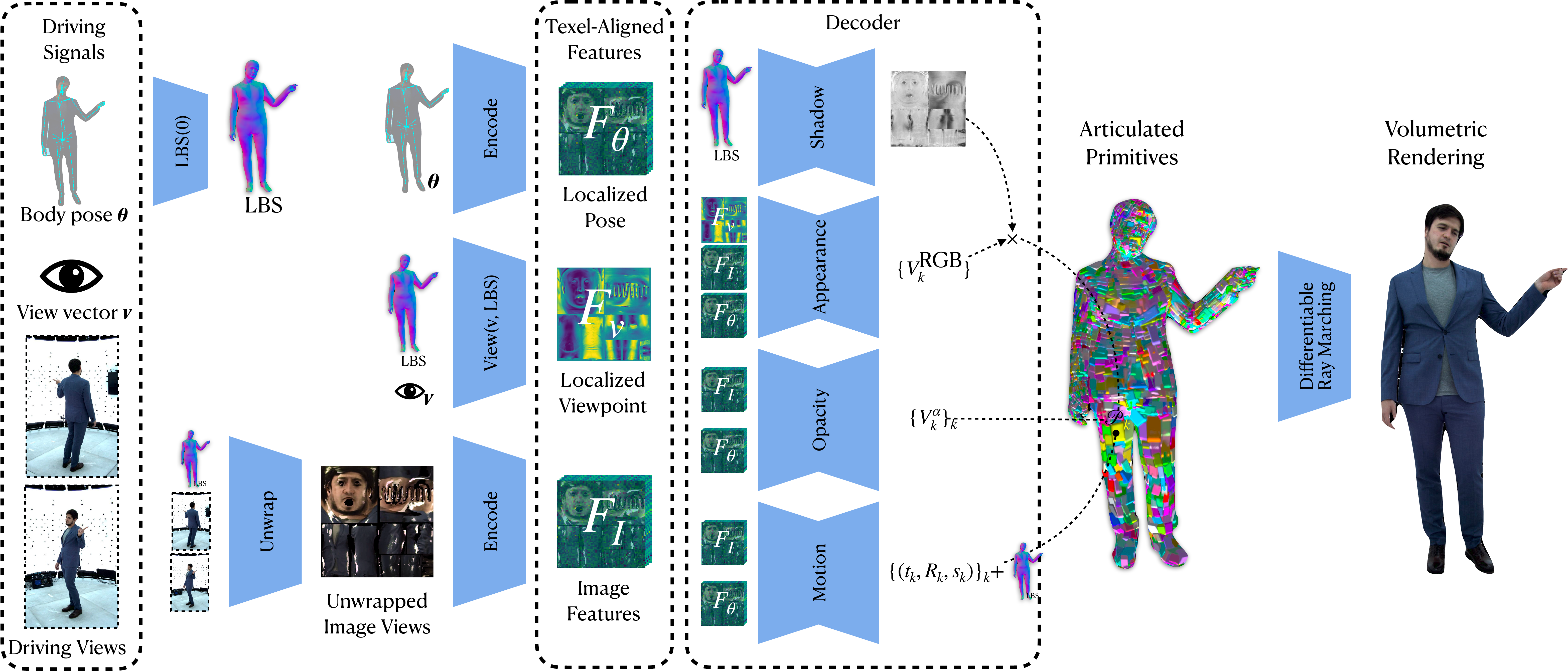}
\caption{\textbf{General overview of the architecture.} 
The core of our full-body model is a encoder-decoder architecture, which takes as input raw images, body pose, facial expression and view direction, and outputs a mixture of volumetric primitives. These are ray marched through to produce a full-body avatar.}
\label{fig:overview}
\end{figure*}

In our experiments, we demonstrate the effectiveness of such hybrid representations 
for full-bodies over the state-of-the-art. 
We also showcase the efficacy of our method by building a complete 
one-way telepresence system, which allows a person to be virtually 
teleported using only a few commodity sensors.

In summary, our contributions are:
\begin{itemize}
    \item We introduce Drivable Volumetric Avatars (DVA): a novel neural 
    representation for animation and free-viewpoint rendering of personalized
    human avatars.
    \item We propose texel-aligned features for DVA: a dense conditioning method that
    leads to better expressiveness and better generalization to novel viewpoints for unseen poses.
    % \item A hybrid volumetric representation for human body geometry and appearance modeling.
    \item We introduce a novel virtual teleportation system
    that uses DVA for one-way photorealistic telepresence.
\end{itemize}
Sample implementation will be made publicly available\footnote{\url{https://github.com/facebookresearch/dva}}.

\section{Related Work}
We first discuss existing representations for modeling dynamic human appearance and geometry. 
We then review existing telepresence systems, with a focus on how 
these systems exploit available driving signals, typically trading 
off robustness for fidelity.

\paragraph{Mesh-Based Avatars}

Textured meshes have been widely used to represent human geometry
and appearance for efficient rendering with modern graphics hardware. 
Parametric human body models can be learned from thousands of scans by deforming a template mesh~\cite{anguelov2005scape,hasler2009statistical,loper2015smpl}. 
These approaches primarily focus on the geometry of minimally clothed human bodies. 
Recent works also model clothing shape variation~\cite{ma2020learning, bhatnagar2019multi}.
Template meshes are also utilized to model the shape and appearance of clothed 
humans from video inputs~\cite{alldieck2018video} or a single image~\cite{alldieck2019tex2shape, weng2019photo}. 
% \ER{In particular, the approaches of \cite{lazova2019360,grigorev2019coordinate} are particularly relevant to our work, as they leverage warping pixels to UV maps to perform novel view synthesis of clothed humans.}
In particular, approaches~\cite{lazova2019360,grigorev2019coordinate} are highly related to our work, 
as they leverage warping pixels to UV maps to perform novel view synthesis of clothed humans.

However, these methods model only static geometry and appearance, failing to produce high-fidelity drivable avatars for novel poses.
Recently~\cite{bagautdinov2021driving} proposed a method to model high-fidelity
drivable avatars from a multi-view capture system by decoding dynamic 
geometry and appearance from disentangled driving signals.
\cite{xiang2021modeling} extends this representation by modeling clothing
explicitly as a separate mesh layer, similarly to ClothCap~\cite{Pons-Moll:Siggraph2017}, 
recovering sharper clothing boundaries.

Despite providing an efficient and effective way to represent dynamic humans, 
these mesh-based approaches require accurate tracking of the underlying geometry of avatars
as a preprocessing step, which significantly limits supported clothing types.
LiveCap~\cite{habermann2019livecap} and its follow-up learning-based method~\cite{habermann2021real} 
simplify the tracking requirement by leveraging silhouette constraints 
while achieving real-time performance. 
However, at the cost of simplification, the fidelity of the resulting avatars are 
not on par with the aforementioned approaches that rely on accurate tracking.
In contrast, our approach, is based on a more flexible volumetric representation, 
simplifying the tracking prerequisites while further improving the fidelity.

\paragraph{Volumetric Avatars}

Recently volumetric representations have been shown effective for modeling 3D humans from a single image~\cite{saito2019pifu, saito2020pifuhd, huang2020arch, li2020monocular, zheng2021pamir}, RGBD inputs~\cite{li2020pifusion, yu2021function4d}, 3D scans~\cite{chibane20ifnet, bhatnagar2020ipnet, Saito:CVPR:2021, palafox2021npms, tiwari21neuralgif}, or multi-view captures~\cite{lombardi2019neural, lombardi2021mixture,peng2021neural, peng2021animatable, liu2021neural, su2021nerf}.

PIFu~\cite{saito2019pifu} and its follow up works~\cite{saito2020pifuhd, huang2020arch, li2020monocular, zheng2021pamir} learn occupancy and texture fields given pixel-aligned image features and 3D coordinates. 
Extending it to RGBD inputs also enables robust avatar creation~\cite{li2020pifusion, yu2021function4d}.
Parametric bodies such as SMPL model~\cite{loper2015smpl} are used to warp 
image-features to a canonical T-pose for modeling animatable avatars from a single image~\cite{huang2020arch, He2021ARCHAC}.
IFNet~\cite{chibane20ifnet} infers implicit functions from partial point clouds by leveraging multi-level features, which is later extended to multi-body parts~\cite{bhatnagar2020ipnet}.
Since volumetric representations support varying topology, modeling animatable clothed avatars
is now possible without explicit surface registration~\cite{Saito:CVPR:2021, palafox2021npms, tiwari21neuralgif}.
While some of these approaches model pose-dependent body geometry, the appearance is either 
ignored or not photo-realistic.

Given multi-view video sequences, Neural Volumes~\cite{lombardi2019neural} models volumetric human heads by decoding radiance in voxels, whose efficiency and fidelity are further improved by Mixture of Volumetric Primitives (MVP)~\cite{lombardi2021mixture}.
Similarly, \cite{POP:ICCV:2021,Ma:CVPR:2021} introduce a collection of articulated primitives to model clothed humans, primarily focusing on geometry.
Neural Body~\cite{peng2021neural} applies differentiable volumetric rendering~\cite{ mildenhall2020nerf} to model articulated human geometry and appearance by diffusing per-vertex latent codes on a SMPL model via sparse 3D convolutions. 
Neural Actor~\cite{liu2021neural} instead projects 3D coordinates on the closest SMPL surface to regress radiance fields. 
A-NeRF~\cite{su2021nerf} and \cite{peng2021animatable} leverages spatial transformations provided by joint articulation for better generalization with unseen poses. 
While the volumetric rendering approaches do not require precise tracking of the surface, they neither run in real-time nor model stochastic nature of clothing deformations as avatars are driven by only pose parameters. 
In contrast, our volumetric avatar runs in real-time, and supports more fine-grained control of the reconstructed avatars such as wrinkles and clothing dynamics.

\paragraph{Drivable Telepresence Systems}
Drivable telepresence systems can be divided into two categories; Non-parametric solutions are holistically modeling a scene and faithfully transmit observed signals as it is~\cite{martin2018lookingood,lawrence2021project, orts2016holoportation}. The other solution is to leverage a category-specific parametric model for driving~\cite{lombardi2018deep, bagautdinov2021driving, xiang2021modeling}. 

Non-parametric telepresence systems, including Holoportation \cite{orts2016holoportation}, and 
Project Starline~\cite{lawrence2021project}, leverage classical 3D reconstruction techniques 
and image-space post-processing for real-time rendering of the scene. 
While these approaches do not require per-user training, its fidelity is bounded by the input signals. 
Thus, occluded regions remain void. 
To alleviate this limitation, LookingGood~\cite{martin2018lookingood} proposes a system to augment real-time performance capture systems with 2D neural rendering. A neural network takes imperfect rendering of captured geometry and appearance, and jointly performs completion, super-resolution, and denoising in real-time. However, we argue that such a image-based solution lacks strong structural prior of human avatar, and results in sub-optimal rendering quality when input views are sparse (i.e., temporal flickering and blur in impainted regions). 
Such a 2D neural rendering technique has been recently applied to human body to better incorporate human prior~\cite{prokudin2021, raj2021anr, shysheya2019textured}. 
However, these approaches rely exclusively on pose for driving the avatars, 
and do not capture remaining information such as clothing deformations and dynamics.

On the contrary, parametric approaches build strong subject-specific priors for face~\cite{lombardi2018deep} 
and body~\cite{habermann2021real,bagautdinov2021driving, xiang2021modeling} by training on a large corpus of multi-view data. 
While the advantage of these approaches lies in the robustness to limited input signals 
(e.g., driving from images from VR headsets~\cite{Wei2019Rosetta}), high compression of drivers' state into 
low-dimensional driving signal often leads to lack of expressiveness for driving. 
In this work, our proposed texel-aligned feature representation allows us to leverage 
strong structural prior provided by a parametric model while recovering fine-grained 
details observed from sparse input images as in non-parametric systems.

\section{Method}

\subsection{Overview}
Our goal is to build a photorealistic personalized avatar of a human
that is expressive and faithful to the driving signal, while also being 
robust to our driving setup with a sparse view inputs.
An overview of our approach is provided in Fig.~\ref{fig:overview}.
Our model is an encoder-decoder architecture that takes as input a 
set of sparse multi-view images, body pose and a viewing direction, 
and produces a collection of volumetric primitives on a human body. The inferred volumetric primitives are then rendered with a differentiable ray marching to produce a photorealistic avatar.
Input pose is used to produce coarse geometry articulated by Linear Blend Skinning~\cite{magnenat1988joint,kavan2008geometric} (LBS). This provides the initial positions of the volumetric primitives.
The underlying skinned model is also used to align all the available driving inputs onto \textit{texel-aligned
features}, which include localized pose, viewpoints, and image features.
Those texel-aligned features are then decoded to a volumetric payload that captures high-resolution
local geometry and appearance as well as dynamic correctives to the primitives' 
transformations.
Similarly to~\cite{bagautdinov2021driving}, we use a shadow branch to capture non-local shading effects,
which also operates in the texture space.

\subsection{Background: Mixture of Volumetric Primitives}

At the core of the MVP~\cite{lombardi2021mixture}, the representation
is a set of $K$ dynamically moving volumetric primitives
that jointly parameterize the color and opacity distribution 
of a modeled scene.
This yields a scene representation that, unlike mesh-based ones, 
is not bound to a fixed topology and, compared to regular volumetric grids~\cite{lombardi2019neural}, 
is memory efficient and fast to render, allowing for real-time rendering 
of high-resolution dynamic scenes.

In practice, each primitive 
$\mathcal P_k = \{ t_k, R_k, s_k, V_k^{\text{RGB}}, V_k^{ \alpha } \} $ 
is parameterized by 
a position $t_k \in \mathbb R^3$ in 3D space,
an orientation $R_k \in \text{SO(3)}$, 
a per-axis scale factor $s_k \in \mathbb R^3$, 
an appearance (RGB) payload
$V_k^{\text{rgb}} \in \mathbb R^{3 \times S \times S \times S}$, 
and opacity 
$V_k^{\alpha} \in \mathbb R^{S \times S \times S}$,
where $S$ is the number of voxels along each spatial dimension.

The synthetic image is then obtained through differentiable ray marching,
using a cumulative volumetric rendering scheme of~\cite{lombardi2019neural}.
More specifically, given a pixel $p$ and corresponding
ray $\mathbf r_p(t) = \mathbf o + t \mathbf d_p$ we compute its 
color $I_p^{\text{rgb}}$ as
\begin{align}
    I_p^{\text{rgb}} &= \int _{t_{\text{min}}} ^{t_{\text{max}}} 
    V^{\text{rgb}}(\mathbf r_p(t)) \frac{d T(t)}{dt} dt \; ,  \\
    T(t) &= \int _{t_{\text{min}}} ^{t_{\text{max}}} V ^{\alpha} (\mathbf r_p(t)) dt \; ,
\end{align}
where $ V^{\text{RGB}}, V^{\alpha}$ denote global color and opacity fields 
and are computed by trilinearly interpolating each primitive hit by the ray.

\subsection{Articulated Primitives}
\label{sec:method:articulatred-prims}

\cite{lombardi2021mixture} introduces a model for human faces, 
where volumetric primitives are loosely attached to a guide mesh 
directly regressed by an MLP.
Unlike faces, however, bodies undergo large rigid motions that
are hard to handle robustly with position-based regression.
In order to generalize better to articulated motion, we 
propose to attach primitives to the output mesh of Linear Blend Skinning model:
\begin{align}
    \mathcal M_{\btheta} = \text{LBS} (\btheta, \mathcal M) \; ,
\end{align}
where $\theta$ denotes human pose, $\mathcal M$ a template mesh in 
canonical pose, and $\mathcal M_{\theta}$ is the final mesh geometry after posing.

We then initialize primitive locations by uniformly sampling UV-space, mapping each primitive to the closest texel, and positioning it at the corresponding surface 
point $\hat t _k (\btheta) \in \mathcal{M}_{\btheta}$.
In practice, we use $K = 4096$ primitives, and thus this
procedure produces a $W\times W, W = 64$ grid, where each primitive is 
\textit{aligned} to a specific \textit{texel}. 
The orientation of the primitives is initialized based on the 
local tangent frame $\hat{R} _k (\btheta)$ of the 3D surface point
on the reposed mesh, and the scale $\hat s _k$ of each primitive is
initialized based on the gradient of 3D rest shape w.r.t. the UV-coordinates 
at the corresponding grid point position. 
%
% In practice, this means that the primitives are initialized with a scale 
% $\hat s _k$ that is inversely proportional to distances to their neighbours.

%
Moreover, although the primitives are associated with the articulated 
mesh, in order to allow for larger variations in topology, they
are allowed to deviate:
\begin{align}
    t_k &= \delta t_k + \hat t _k (\btheta), \\
    R_k &= \delta R_k \cdot \hat{R} _k (\btheta), \\
    s_k &= \delta s_k + \hat s _k,
\end{align}
where $\delta t_k, \delta R_k, \delta s_k $ are primitive 
\textit{correctives} that are produced by the motion branch of our decoder. 

% Fig.~\ref{fig:mvp} depicts primitives at initialization and once 
% the optimization has converged for one of the subjects in our dataset, 
% observe how primitives are sampled more densely for parts of the 
% human body that might require that, such as face or hands.

\subsection{Texel-Aligned Features}
\label{sec:method:texel-features}
\label{sec:method:view-cond}

In this section, we describe our novel dense representation which
allows us to fully exploit available driving signals.
As discussed earlier, our primitives are aligned into a $W \times W$
2D grid, where each primitive is assigned to a specific texel 
on the UV-map.
In order to condition each of the primitives only on the relevant 
information, we propose to use the same spatial prior to align 
all the input signals to the corresponding structure.

\paragraph{Body Pose} For body pose $\btheta$, we employ 
location-specific encodings 
similar to \cite{bagautdinov2021driving}:
we use skinning weights to limit the spatial extent 
of each pose parameter, project the resulting masked features
to the UV-space at the same $W \times W$ resolution, 
and then apply a dimensionality reducing projection, 
implemented as a 1x1 convolution, to get $F_{\btheta}$. 
Such localized representation helps
limit overfitting, and reduces the tendency to learn
spurious long-range correlations which might be present
in the training data as shown in prior works~\cite{Saito:CVPR:2021, bagautdinov2021driving}. 

\paragraph{Images} Conditioning the model only on pose is insufficient
for faithful driving, as it does not contain 
all the information required to explain the entire appearance of 
a clothed human in motion, such as stochastic clothing state and dynamics.
To this end, we propose to use available image evidence,
by projecting it to a common texture space.
Namely, for each available input view, we back-project image pixels corresponding to all visible vertices of our posed LBS template to UV-domain. 
Once all visual evidence has been mapped to a common UV-domain, 
we average it across all the available views to get 
a multi-view texture.
%
% This results in image features that can capture information
% that may not be explained by pose.  
%
We then compress the resulting high-resolution texture to the 
same resolution as $F_{\btheta}$ with a convolutional encoder
to obtain $F_{\bI}$.
Note that, unlike the existing 
works~\cite{bagautdinov2021driving,lombardi2018deep}, which
assume all the remaining information is encoded into a global low-dimensional code, our representation preserves spatial 
structure of the signal and is significantly more expressive,
thus allowing to better capture deformations present in the input signals (see Fig.4).

\paragraph{Viewpoint} The existing work on modeling view-dependent appearance of human body~\cite{peng2021neural,peng2021animatable,liu2021neural, bagautdinov2021driving} represents viewpoint globally for the 
entire body, as a relative camera position w.r.t. the root joint.
However, this representation is not explicitly taking 
into account articulation, and the model is forced to learn complex
interactions between pose-dependent deformations and viewpoint
from limited data, which leads to overfitting (see Fig.~\ref{fig:viewcond}). 
To address this, we encode the camera position in the  \textit{local} coordinate
frame of each primitive.
Specifically, given a view-direction $\bv \in \mR^{3}$ and the
posed template mesh $\mathcal M _{\theta}$, we compute 
per-triangle normals $\bn_t$ and use them to express camera 
coordinates relatively to the local tangent plane of each 
triangle as
\begin{align}
    v_t = \bv \cdot \bn_t.
\end{align}
Then, we warp this quantity to UV space, and sample it at $W \times W$
resolution to get texel-aligned viewpoint features $F_{\bv}$.

\subsection{Architecture and Training Details}

Given texel-aligned features, our decoder produces the payload of the volumetric primitives (Fig.~\ref{fig:overview}).
In practice, we employ three independent branches, each
being a sequence of 2D transposed convolutions with untied 
biases that preserve spatial alignment.
The motion branch is conditioned on $(F_{\btheta}, F_{\bI})$, 
and produces transformation correctives 
$\{(\delta t_k, \delta R_k, \delta s_k)\}_k \in \mR^{9 \times W \times W}$, 
which are then applied to the initial locations of 
articulated primitives.
The opacity branch is conditioned on $(F_{\btheta}, F_{\bI})$ and produces 
a slab $\{V^{\alpha}_k\}_k \in \mR^{S \times W\cdot S \times W \cdot S}$,
where $S = 16$ is the number of voxels along a spatial dimension.
The appearance branch is conditioned on $(F_{\btheta}, F_{\bI}, F_{\bv})$,
and produces a slab $\{V^{\text{rgb}}_k\}_k \in \mR^{3 \times S \times W \cdot S \times W \cdot S }$.
Additionally, to capture long-range pose-dependent effects, we employ a shadow branch in~\cite{bagautdinov2021driving} by replacing the output channel of the last convolution layer to match with the ${V^{\text{rgb}}_k}$ for multiplication.
 
% and multiply the appearance slab ${V^{\text{rgb}}_k}$ by its output.

We use the following composite loss to train all our models:
\begin{equation}
\mathcal{L} = 
\lambda_{\text{rgb}} \mathcal{L}_{\text{rgb}} + 
\lambda_{\text{vgg}} \mathcal{L}_{\text{vgg}} + 
\lambda_{\text{m}} \mathcal{L}_{\text{m}} +
\lambda_{\text{vol}} \mathcal{L}_{\text{vol}} \;,
\end{equation}
where $\mathcal{L}_{\text{rgb}}$ is the MSE image loss,
$\mathcal{L}_{\text{vgg}}$ is the perceptual VGG-loss,
$\mathcal{L}_{\text{m}}$ is the MAE segmentation mask loss,
and $\mathcal{L}_{\text{vol}}$ is the volume prior loss~\cite{lombardi2021mixture}
that encourages primitives to be as small as possible.
%
% In all our experiments, we use Adam~\cite{} optimizer with 
% learning rate $0.001$ to minimize $\mathcal{L}$.
%
Empirically, we found that it is important to train the model 
in two stages to ensure robustness with respect to the quality 
of tracking and LBS model.
Namely, for the first $N = 1000$ iterations we condition our model on 
all available training views (for our data ~160 cameras), 
and then continue training on a sparse signal computed from
3 randomly sampled view.
Intuitively, this helps our model pick up useful signals by gradually 
shifting from easy samples to harder ones in the spirit of curriculum
learning~\cite{BengioLCW09}. 
% Intuitively, this helps model to start picking up cues from 
% conditioning signal instead of ignoring it completely, 
% yet still eventually has to learn meaningful feature encoding
% so as to minimize the training loss.

\section{Experiments}

In this section, we report our experimental findings, ablate
different components of our method (DVA), and showcase a 
teleportation system that uses DVA to create a one-way
photorealistic telepresence experience.

\subsection{Datasets}
We report most of our results on data acquired with a setup
similar to~\cite{bagautdinov2021driving, xiang2021modeling}:
a multi-view dome-shaped rig with ~160 high-resolution (4K)
synchronized cameras. We collect data for three different identities,
including one with challenging multi-layer clothing (a suit), 
run LBS-tracking pipeline to obtain ground truth poses, and then 
use roughly 1000 frames in various poses for training our models.
Additionally, we evaluate the performance of our articulated volumetric 
representation for human bodies on a public dataset
ZJU-MoCAP~\cite{peng2021neural}\footnote{No facial meshes were created for the individuals in ZJU-MoCAP, and the dataset was not used to identify individuals}.

\subsection{Ablation Study}

% \begin{table}[h]
%     \begin{tabular} {c|c|c}
%         \hline
%         Method & {Supervision} & PSNR \\ 
%         \hline
%         OURS & 3D+2D & 31.011 \\ 
%          & 2D & 30.920  \\ 
%          \hline
%         FBCA & 3D+2D & 30.519 \\ 
%          & 2D &  30.491 \\ 
%     \end{tabular} 
%     \caption{Importance of 3D supervision.}
%     \label{tab:3Dvs2D}   
% \end{table}
% \paragraph{Importance of 3D supervision} In this experiment, we evaluate if, when available, 3D supervision can be used to improve the accuracy of our model. 
% %
% To this end, we modify our LBS articulation to also include 
% a geometry branch, producing dynamically changing correctives to the
% LBS template, and supervise it with dense correspondences from 
% tracked meshes.

% \ER{Metrics and quality is very much comparable for both OURS and FBCA with and without 3D supervision on this experiment. 
% Not sure on what should we conclude from this, other than LBS template + tracking is really really good.
% I think the experiment on ZJU is more informative: it shows it's not the case when LBS is bad, as MVP can work without supervision whereas FBCA struggles. 
% So I would remove this ablation completely, and just show that we outperform FBCA on ZJU.}

\begin{figure}[ht!]
\begin{overpic}[width=0.48\textwidth]{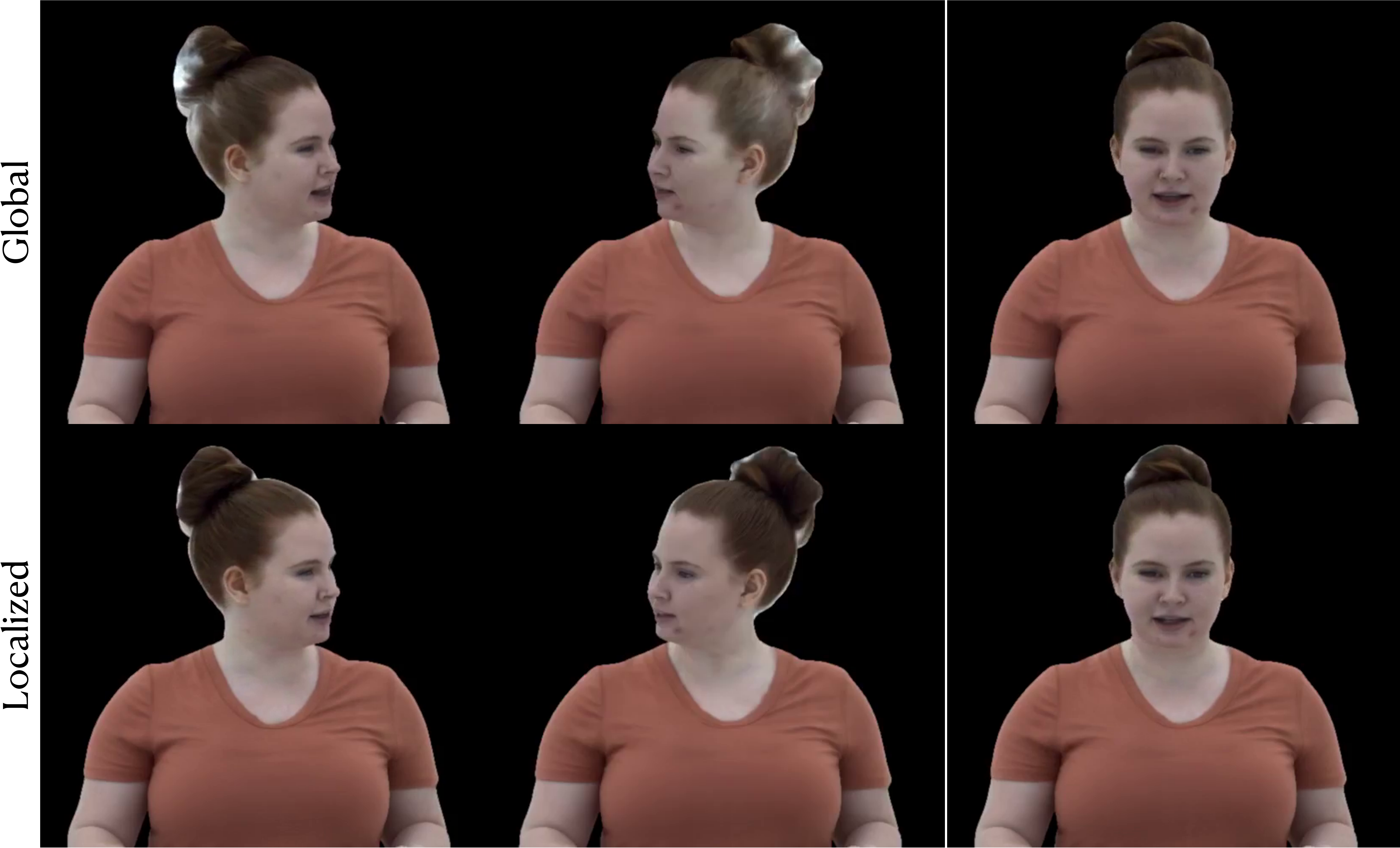}
\put(32,-3.5){{Test}}
\put(80,-3.5){{Train}}
\end{overpic}
\caption{\textbf{Effects of view conditioning.} 
Localized view conditioning leads to better generalization
on unseen combinations of poses and viewpoints.}
\label{fig:viewcond}
\end{figure}

\paragraph{View Conditioning.} In this experiment, we demonstrate
the effectiveness of the localized view conditioning described
in Section~\ref{sec:method:texel-features}. 
In Fig.~\ref{fig:viewcond}, we show qualitative performance of
a version of our model trained with local view conditioning and the
instance trained with the global one~\cite{bagautdinov2021driving}.
Our localized view conditioning leads
to plausible view-dependent appearance with unseen poses, whereas the global view conditioning
suffers from significant visual artifacts due to overfitting.

% \textbf{Driving signal ablation / comparison to NeuralActor}
% \er{this should ablate our input representations (RGB and PiFU if we decide to go for it, benefits are unclear still) vs NeuralActor}

\begin{figure}[ht!]

\begin{overpic}[width=0.48\textwidth]{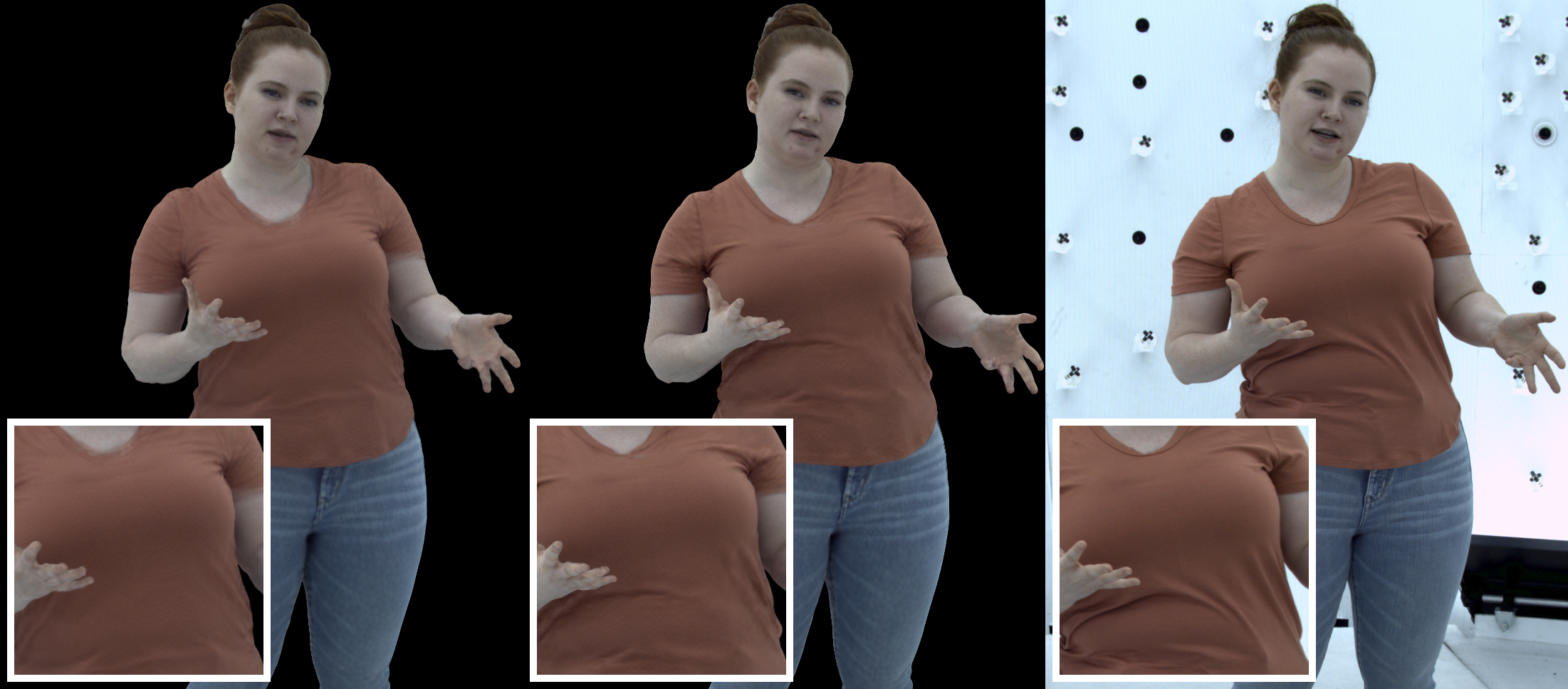}
\put(7,-3.5){{Bottleneck}}
\put(40,-3.5){{Texel-Aligned}}
\put(73,-3.5){{Ground Truth}}
\end{overpic}

\caption{\textbf{Effects of texel-aligned features.} 
Expressive texel-aligned features allow our model 
to generalize better to challenging unseen 
clothing states.}
\label{fig:texel}
\end{figure}

\paragraph{Texel-Aligned Features.} In Fig.~\ref{fig:texel}, we 
provide qualitative comparison of two different instances of our
method: one that uses texel-aligned features, and one that relies on a
bottleneck representation akin to~\cite{bagautdinov2021driving}.
The instance with texel-aligned features demonstrates significantly better preservation of high-frequency details 
with respect to the ground truth.

% \textbf{Localized pose conditioning vs global pose conditioning.}
% Ablation is running, not sure we need it though in light of the fact that FBCA is also doing that and it ablated its effectiveness?

\subsection{Novel View Synthesis}
In order to evaluate the effectiveness of our articulated volumetric 
representation with respect to existing methods, we provide quantitative 
and qualitative comparisons on ZJU-MoCap~\cite{peng2021neural} dataset.
The goal of this challenging benchmark is to produce a photorealistic novel view synthesis (NVS) of a clothed human in motion, while training only from 4 views.

\begin{table}[h]
    \caption{Quantitative results (PSNR) for NVS on ZJU-MoCap.}
    \begin{tabular} {c|c| c}
        \hline
        Method & S386 & S387 \\ 
        \hline
        FBCA & 32.123 & 27.886 \\ 
        NeuralBody  & 33.196  & 28.640\\ 
        OURS & \textbf{35.414} & \textbf{30.512}  \\ 
        \hline
    \end{tabular} 
    
    \label{tab:3Dvs2D}   
    
\end{table}

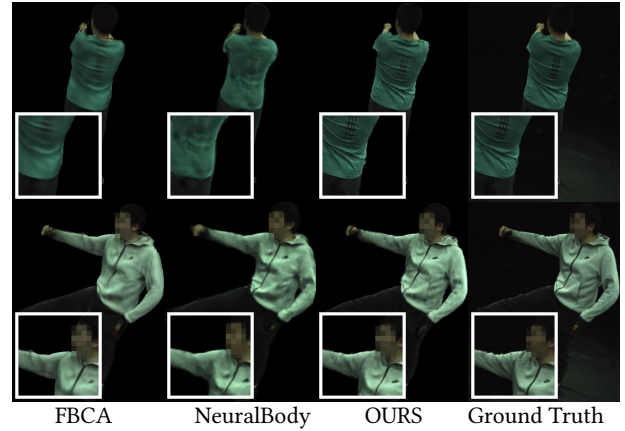
\begin{figure}[ht!]
\centering
\begin{overpic}[width=0.45\textwidth]{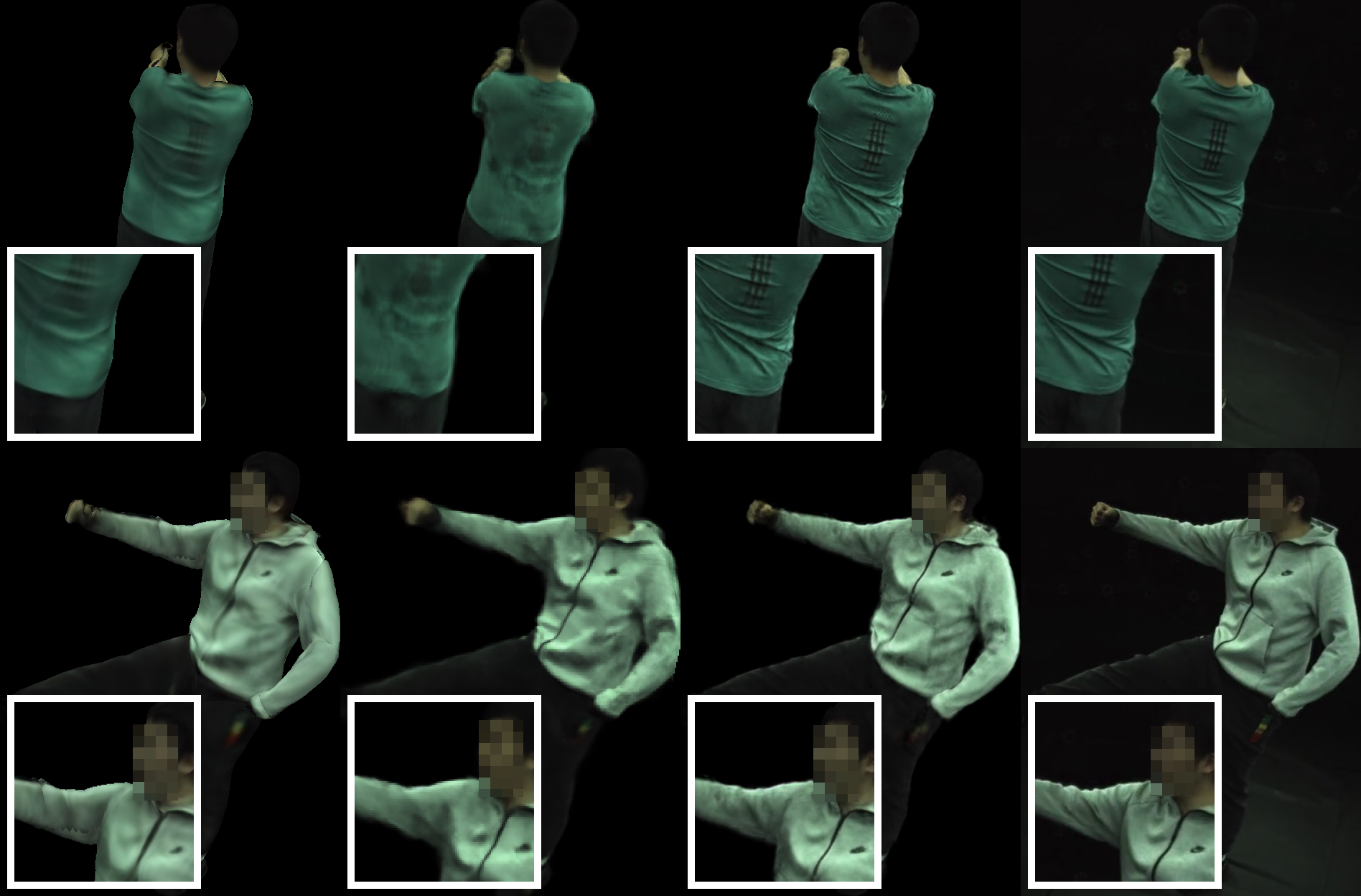}
\put(7,-3.5){{FBCA}}
\put(30,-3.5){{NeuralBody}}
\put(58,-3.5){{OURS}}
\put(75,-3.5){{Ground Truth}}
\end{overpic}
\caption{\textbf{Novel View Synthesis.} We compare our method to state-of-the-art for NVS on ZJU-MoCap.
Despite not being explicitly tailored to be trained with sparse supervision, our method outperforms competitors.
Real faces and their reconstructions are blurred for anonymity.}
\label{fig:3Dvs2D}
\end{figure}

We compare to Full-Body Codec Avatars (FBCA)~\cite{bagautdinov2021driving}
and Neural Body~\cite{peng2021neural}, 
which represent the state-of-the-art among respectively mesh-based
and volumetric approaches.
In Tab.~\ref{tab:3Dvs2D}, we provide quantitative results for novel view
synthesis in terms of PSNR for two subjects, which suggest that our
method provides significant improvements over baselines.
In Fig.~\ref{fig:3Dvs2D}, we provide a qualitative comparison; our 
method produces sharper reconstructions and less artifacts 
than both of the baselines.
Interestingly, the mesh-based FBCA is performing significantly worse
in settings without direct mesh supervision by precise multi-view stereo and tracking (the only source of geometry supervision in ZJU-MoCap dataset is silhouette and image losses through
differentiable rendering). In constrast, our volumetric approach
is able to learn more accurate underlying geometry with only image-based supervision due to the flexibility of volumetric representation.
%
% We believe that this provides evidence that hybrid volumetric 
% approaches are capable of learning high-quality without
% direct 3D supervision.
%
Please refer to the supplemental video for more detailed visual comparison.
% We provide a visual comparison in the supplemental video.
% \ER{Compare with methods that are tailored to novel view synthesis on ZJU Mocap, assuming SMPLX tracked poses in input and no 3D supervision.
% This shows that our methods works without 3D supervison, unlike FBCA.
% Results on other 3 performer are being generated.}

\subsection{Driving Results}

We compare our approach to two different kinds of drivable telepresence systems: 
a mesh-based model (FBCA), and image-space model (LookingGood),
both trained on our high-quality multi-view captures.
Note that, in practice we use our own re-implementation of LookingGood,
which uses LBS tracking instead of raw depth maps.

\begin{table}[ht!]
    \caption{Quantitative results (PSNR) on unseen motion. Please refer to supplementary video for more results.}
    \begin{tabular} {c|c|c|c|c}
        \hline
        \multirow{2}{*}{Method} & \multirow{2}{*}{Views} & \multicolumn{3}{c}{Test view} \\
         &  & \texttt{front} & \texttt{back} & \texttt{avg} \\
        \hline
        FBCA & None & 30.571 &	30.656 &	30.613 \\ 
        \hline
        \multirow{2}{*}{LookingGood}    & 2 & 33.442 &	26.059 & 29.750  \\
            & 3 & 33.256	& 32.453 &	32.854  \\
        \hline
        \multirow{2}{*}{OURS}  & 2 & 33.615 &	33.203	& 33.409 \\ 
          & 3 & \textbf{33.617} & \textbf{33.838} & \textbf{33.728} \\ \hline
    \end{tabular} 
    
    \label{tab:quant-driving}   
\end{table}

Results of quantitative evaluation are provided in Table~\ref{tab:quant-driving}. Models are evaluated on two different views, 
\texttt{front}-facing and \texttt{back}-facing.
To evaluate robustness to the sparsity of input views, we consider two different settings for the approaches that utilize input images as driving signal: 
one where we provide only conditioning from 2 front facing cameras, and a less challenging one where we provide 3 
uniformly sampled conditioning views.
Our model outperforms both of the baselines, and is more robust 
to missing information compared to LookingGood, in particular
on settings with severe sensory deprivation.
Qualitative results are provided in Fig.~\ref{fig:qual-driving}.
We provide additional comparisons on dynamic sequences in supplementary video.

\begin{figure*}[ht!]
\hspace{2pt}
\centering
\begin{overpic}[width=\textwidth]{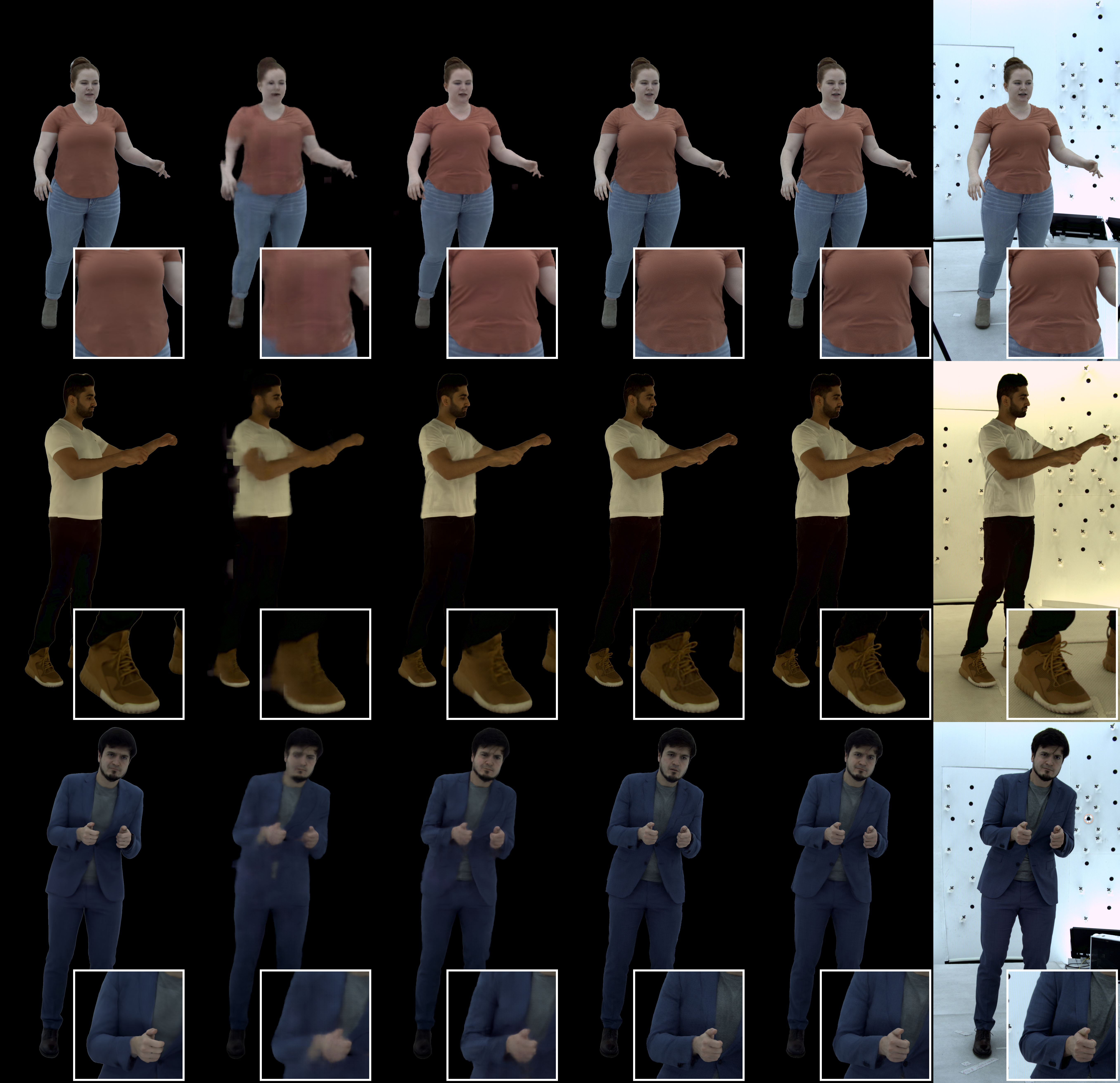}
\put(6,-1.7){{FBCA}}
\put(18,-1.7){{LookingGood (\texttt{V=2})}}
\put(35,-1.7){{LookingGood (\texttt{V=3})}}
\put(53,-1.7){{OURS (\texttt{V=2})}}
\put(69,-1.7){{OURS (\texttt{V=3})}}
\put(86,-1.7){{Ground truth}}
\end{overpic}
\caption{\textbf{Qualitative results: Driving.} We compare our method to state-of-the-art approaches for drivable avatars on unseen sequences from our dataset. Best seen in supplemental video. $V$ is the number of view inputs.}
\label{fig:qual-driving}
\end{figure*}
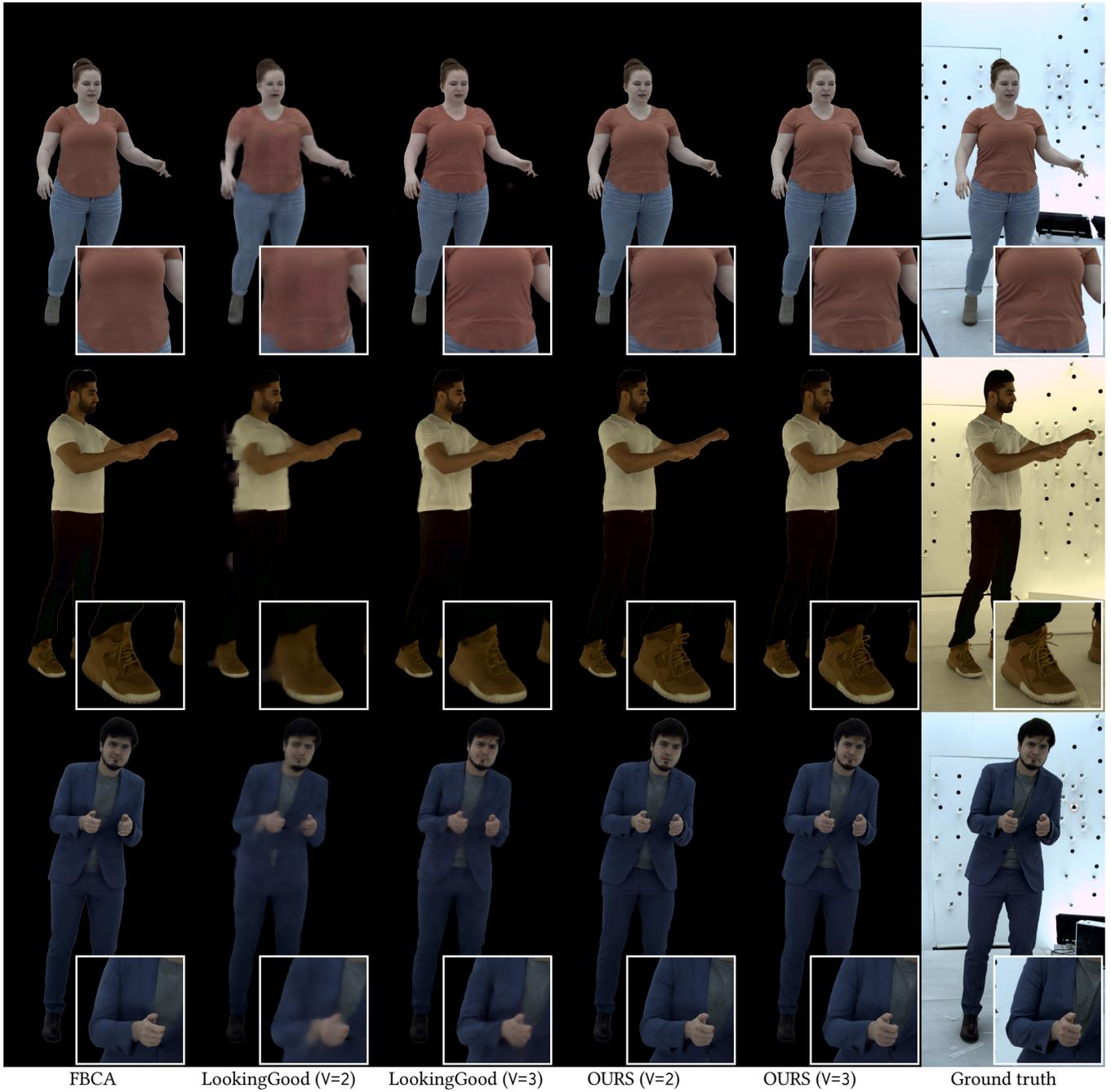

\subsection{Teleportation}

% \begin{figure*}[ht!]
% \centering
% \includegraphics[width=\textwidth]{figs/demo.png}
% \caption{\textbf{Social Teleportation.} We demonstrate our method for one-way virtual teleporation. Given input images and their estimated 3D pose and facial expression (a), we backproject images to uv coordinates (c), and use use our encoder-decoder architecture to obtain a full-body avatar (d), which closely resembles ground truth (d).}
% \label{fig:teleportation}
% \end{figure*}
We also demonstrate the versatility of our method and show that our reconstructed avatars can be driven outside the capture system used to generate training data without losing details.
% it 
% can generalize to driving signals that are extracted
% from a much lighter capture system than the complex capture system used to generate training. 
% capture system used to generate training data., while still preserving 
% details, by relying on texel-aligned features.
%
Our setup consists of 8 synchronized and calibrated Microsoft Azure Kinects cameras~\footnote{\url{https://azure.microsoft.com/en-us/services/kinect-dk/}}, uniformly placed in a circle of 4.5-meter diameter.
To obtain body poses (LBS parameters), we fit a pre-built personalized LBS body model to a sequence of detected and triangularized keypoints from RGB images~\cite{Wei:2016:Convolutional}, as well as meshes obtained by fusing multiview point clouds~\cite{yu2021function4d}.
To obtain texel-aligned features, we simply apply texture unwrapping as in our data processing for the capture dome.
%To obtain texel-aligned features, we run an off-the-shelf model~\cite{saito2020pifuhd}\ShS{TODO(Edo): Check} to generate dense surface normal maps, that are later unwrapped to texture space.
%\ER{ I actually haven't noticed a great difference in RGB vs pseudo normals, and the results in the video are from RGB}\ShS{updated}
%
Even though these driving signals are obtained from unseen sequences under 
different sensor modality and pre-processing, 
we can still faithfully animate our avatars. 
Figure~\ref{fig:teaser} shows that the animated avatars preserve local details 
such as wrinkles on the clothes without noticeable artifacts in appearance.
% without noticeable artifacts in appearance and preserving local details such as wrinkles on the clothes,
% %
% as shown in Figure~\ref{fig:teaser}. 
%\color[red]{SW: more result figures?}
% 
For more results, please refer to the supplementary video.

\section{Conclusion}
We introduced Drivable Volumetric Avatars, a novel method 
for building expressive fully articulated avatars and faithfully driving it from sparse view inputs.
Our approach combines the robustness of parametric 
models by incorporating a strong articulated volumetric prior, 
and the expressiveness of non-parametric models by leveraging 
texel-aligned features.
We demonstrated the efficacy of our method on novel view synthesis and
driving scenarios, and showcased a one-way teleportation system based on our approach to create a photorealistic 
telepresence experience.
Some of the main limitations of our work originate
from our reliance on LBS tracking: our model
still requires cumbersome skeleton 
tracking as a pre-processing step, and cannot handle very loose clothing that significantly deviates from the guide LBS mesh.
A potential avenue for future work is extending the
model to multi-identity settings, multiple outfits, and driving from a head-mounted capture device for a two-way telepresence system.

\clearpage

\bibliographystyle{ACM-Reference-Format}
\bibliography{bibliography}

\end{document}